\documentclass[letterpaper]{article} 
\usepackage{aaai23}
\usepackage{times}  
\usepackage{helvet}  
\usepackage{courier}  
\usepackage[hyphens]{url}  
\usepackage{graphicx} 
\usepackage{natbib}  
\usepackage{caption} 
\usepackage{algorithm}
\usepackage{algorithmic}

\usepackage{hyperref}

\nocopyright

\usepackage{newfloat}
\usepackage{listings}
\DeclareCaptionStyle{ruled}{labelfont=normalfont,labelsep=colon,strut=off} 
\floatstyle{ruled}
\newfloat{listing}{tb}{lst}{}
\floatname{listing}{Listing}
\title{DyRRen: A Dynamic Retriever-Reranker-Generator Model for\\Numerical Reasoning over Tabular and Textual Data}
\author{
    Xiao Li,
    Yin Zhu,
    Sichen Liu,
    Jiangzhou Ju,
    Yuzhong Qu,
    Gong Cheng\thanks{Contact Author}
}
\affiliations{
    State Key Laboratory for Novel Software Technology, Nanjing University, Nanjing, China\\
    \{xiaoli.nju, yinzhu, sichenliu, jujiangzhou\}@smail.nju.edu.cn, \{yzqu,gcheng\}@nju.edu.cn
}

\usepackage{bibentry}

\usepackage{amsthm,amsmath,amssymb}
\usepackage{color}
\usepackage{mathrsfs}

\begin{document}

\maketitle

\begin{abstract}
Numerical reasoning over hybrid data containing tables and long texts has recently received research attention from the AI community. To generate an executable reasoning program consisting of math and table operations to answer a question, state-of-the-art methods use a retriever-generator pipeline. However, their retrieval results are static, while different generation steps may rely on different sentences. To attend to the retrieved information that is relevant to each generation step, in this paper, we propose DyRRen, an extended retriever-reranker-generator framework where each generation step is enhanced by a dynamic reranking of retrieved sentences. It outperforms existing baselines on the FinQA dataset.

\end{abstract}

\section{Introduction}
Numerical reasoning has drawn much attention in machine reading comprehension tasks. Previous datasets on numerical reasoning like DROP~\cite{drop} mainly focus on simple calculations and comparisons. Recently in this field, more complicated arithmetic expression generation takes an important part, especially in the context of financial reports. Recognized datasets like FinQA~\cite{finqa} and TAT-QA~\cite{tat-qa} involve hybrid question answering~(QA), i.e., QA over information represented in heterogenous forms including tables and texts.

\begin{figure}[t]
    \centering
    \includegraphics[width=\linewidth]{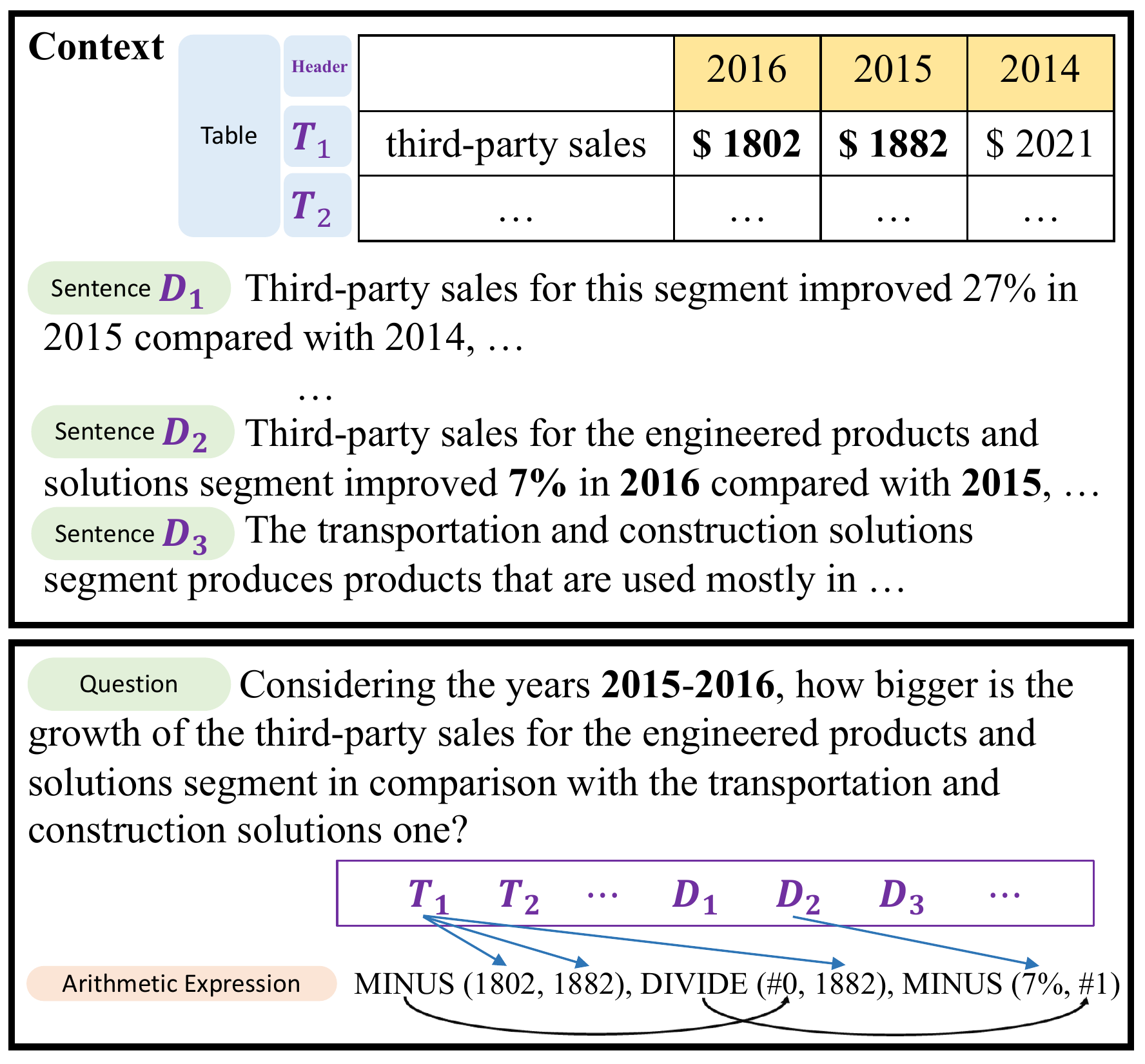}
    \caption{A question sampled from the FinQA dataset.}
    \label{fig:example}
\end{figure}

To acquire an answer by numerical reasoning, a standard method is to generate an arithmetic expression from structured tables and unstructured sentences. For instance, Figure \ref{fig:example} shows an example in FinQA. The context includes a table describing sales records in different years, and several sentences stating relevant financial information. According to the question, one needs to determine arithmetic operators $\mathtt{MINUS}$ and $\mathtt{DIVIDE}$, and select their arguments from either tables or sentences to generate the correct arithmetic expression and calculate the answer. In some cases, an argument of an arithmetic expression~(e.g., \texttt{\#0} in Figure~\ref{fig:example}) could be the result of some former expression.

\subsection{Existing Methods and Limitations}
Intuitively, the generation of an arithmetic expression can be divided into two modules: Retriever and Generator. The retriever filters the most relevant information as the input to the generator which then predicts expressions using operators predefined in a domain specific language (DSL) and numbers extracted from the input.
FinQANet~\cite{finqa} is one such state-of-the-art method. Its retriever converts a table into sentences. Then a BERT~\cite{bert} binary classifier is applied to identify relevant sentences, based on which an encoder-decoder model is implemented to generate an arithmetic expression. Encoders can be BERT or RoBERTa~\cite{roberta} which are both encoder-only models. The hidden layer output of an LSTM~\cite{lstm} is used to predict expression tokens.

We observe the following limitations of FinQANet.

First, in FinQANet's generator, retrieved sentences are concatenated into a long sequence and inputted into the encoder, from which different numbers are extracted at different generation steps. Existing models have difficulty in locating the correct number from a long sequence. For example, when generating the argument ``7\%" in Figure~\ref{fig:example}, the generator should pay exclusive attention to the unique retrieved sentence~$D_2$ which contains ``7\%".
Therefore, it would be helpful to design a mechanism that can dynamically adjust the attention of the generator at generation time.

Second, there is plenty of room to improve FinQANet's retriever which is a simple binary classifier. Indeed, it could not fully exploit the relationships between question and context. For example, in the question in Figure~\ref{fig:example}, the retriever should focus on particular entities such as ``2015", ``2016", and ``third-party sales", which requires intensive interaction at the token level. 
Moreover, it could not directly capture the differences between positive and negative sentences.

\subsection{Our Approach}
\begin{figure}[t]
    \centering
    \includegraphics[width=\linewidth]{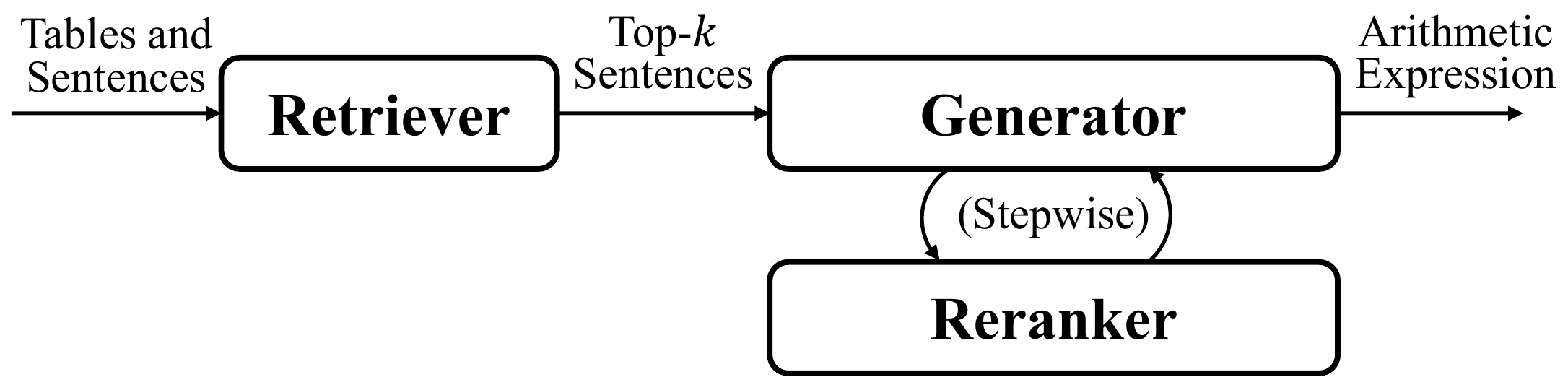}
    \caption{Overview of DyRRen.}
    \label{fig:mini_model}
\end{figure}

We overcome the above limitations with our novel framework DyRRen, short for \textbf{Dy}namic \textbf{R}etriever-\textbf{R}eranker-g\textbf{en}erator. Figure~\ref{fig:mini_model} sketches our framework. DyRRen
switches its attention to different sentences at different generation steps via a novel dynamic reranking mechanism.

\subsubsection{Generator}
In the generator, we focus on scoring the input tokens and looking for the next token most likely to appear in the generated expression. This module is implemented by an LSTM with the attention mechanism.
To improve the accuracy of the generator, we incorporate a reranker to dynamically pick relevant sentences.

\subsubsection{Reranker}
At each step of generation, we score and rerank retrieved sentences in order to find one that contains the target number. Specifically, we take advantage of the relationships among the decoder output of the generator, formerly generated expression, and retrieved sentences to dynamically rerank sentences at each step to enhance the generator.

\subsubsection{Retriever}
Dense retrieval is used to replace the simple BERT binary classifier. To improve retrieval performance, we implement meticulous interaction between each question-sentence pair and employ a set of useful mechanisms such as token-level late interaction. The loss function is also tailored to the scenario in which the retrieval space contains multiple positive and negative sentences.

To summarize, our contributions include
\begin{itemize}
    \item A novel retriever-reranker-generator framework for solving numerical reasoning over tabular and textual data, especially using a reranker which can dynamically locate target information at different generation steps, and
    \item A thoughtfully designed retriever in which we strengthen token-level and pairwise interaction.
\end{itemize}

\subsection{Code}
Our code is available on GitHub: \url{https://github.com/nju-websoft/DyRRen}.

\subsection{Outline}
We elaborate our approach in Section \ref{sec:approach}, present experiments in Section \ref{sec:experiments}, discuss related work in Section \ref{sec:related_work} and conclude our work in Section \ref{sec:conclusion}.
\section{Approach}\label{sec:approach}
In the task of numerical reasoning over tabular and textual data, a question $Q$ is given together with a structured table $T$ which consists of $r$ rows $\{T_1,T_2,\cdots,T_r\}$ and a set of $n$ unstructured sentences $\mathbb{D}=\{D_1,D_2,\cdots,D_n\}$. The goal is to generate an arithmetic expression $G=(\text{op}_1(\text{args}_{11},\text{args}_{12}),\text{op}_2(\text{args}_{21},\text{args}_{22}),...)$, where $\text{op}_i$ is a binary operator such as $\mathtt{DIVIDE}$ defined in a domain specific language (DSL) and $\text{args}_{ij}$ is either a constant 
token, a memory token indicating the result of some previous operation, or a span\footnote{It could be a number or a row header. For clarity, we will refer to it as a numeric argument/token in the following.} that appears in $Q$, $T$, or $\mathbb{D}$.

We extend the existing retriever-generator framework to a retriever-ranker-generator framework.
We first use the retriever to retrieve the most relevant sentences to the question (Section~\ref{sec:retriever}), and then use the retrieved sentences to generate arithmetic expressions (Section~\ref{sec:generator}) which is enhanced by a dynamic reranker of sentences~(Section~\ref{sec:reranker}).

\begin{figure*}[ht]
    \centering
    \includegraphics[width=0.9\linewidth]{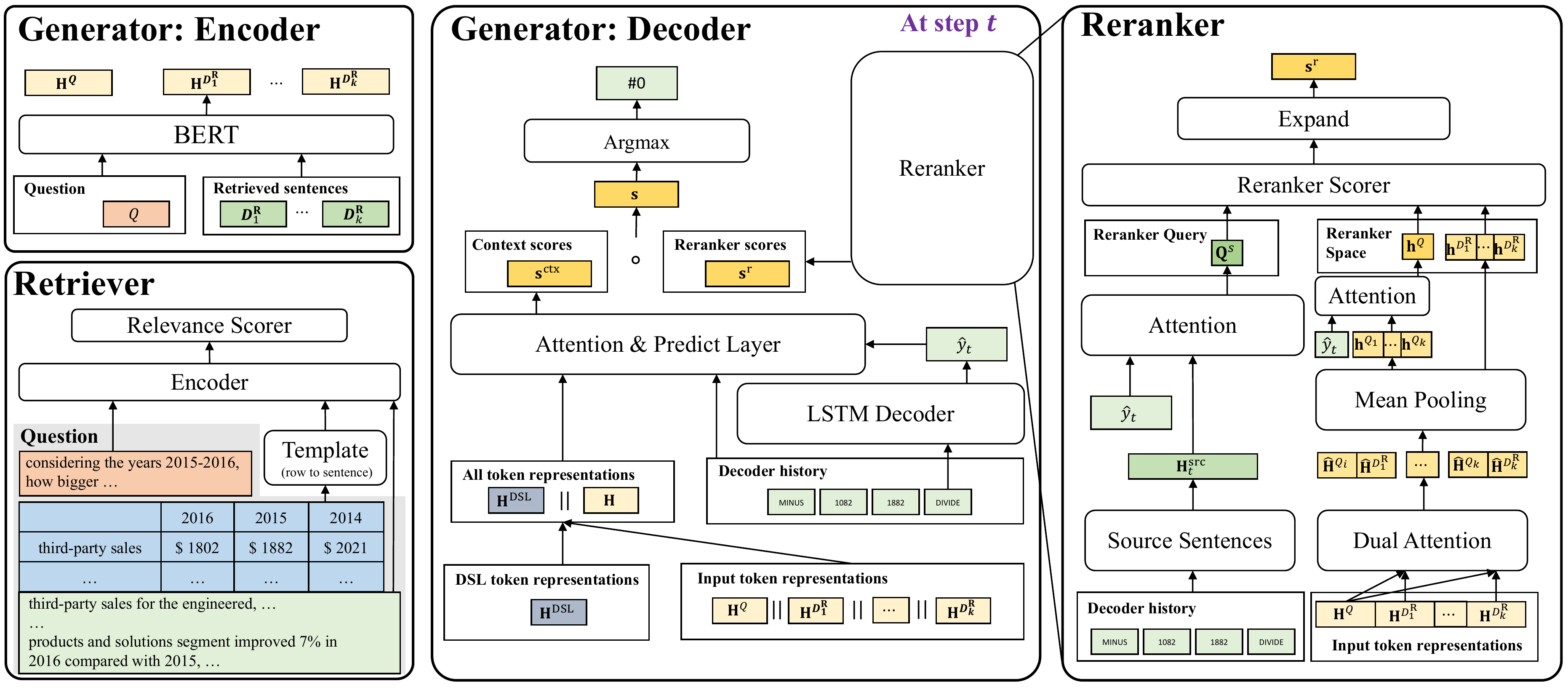}
    \caption{Implementation of DyRRen.}
    \label{fig:generator_model}
\end{figure*}

\subsection{Retriever}\label{sec:retriever}
Given the question $Q$, table $T$ and sentences $\mathbb{D}$, our retriever aims to retrieve supporting facts from $T$ and $\mathbb{D}$ to answer $Q$. The retriever is shown bottom left in Figure~\ref{fig:generator_model}.
\subsubsection{Encoder}
In the retriever, we process tabular data in the same way as FinQANet~\cite{finqa}, converting each row of the tabular data into a sentence using templates.
Specifically, we convert each cell into ``the \textbf{column name} of \textbf{row name} is \textbf{cell value} ;" and then we concatenate the converted cell sentences of a row into a row sentence. For example, the second row of the table in Figure~\ref{fig:example} will be converted to ``the \textbf{third-party sales} of \textbf{2016} is \textbf{\$ 1802}; the \textbf{third-party sales} of \textbf{2015} is \textbf{\$ 1882}; ...".

Recall that the retriever of FinQANet uses BERT encoder to obtain the representations of sentences (including sentences converted from tables). Then it uses a binary classifier to determine whether each sentence is relevant. Different from FinQANet's retriever, we use dense retrieval to improve the performance of the retriever.

Specifically, by using templates to convert each row of $T$ into a sentence, we obtain a set of sentences
\begin{equation}
    \mathbb{D}'=\{D_1,D_2,\cdots,D_n\}\cup\{D_{n+1},D_{n+2},\cdots,D_{n+r}\}
\end{equation}
where $\{D_{n+1},D_{n+2},\cdots,D_{n+r}\}$ are converted from rows.

We use BERT to encode question $Q$ and each sentence $D\in\mathbb{D}'$ into $\mathbf{H}^Q$ and $\mathbf{H}^D$ respectively.
More details about our encoder will be introduced in Section~\ref{sec:generator}. Inspired by ColBERT~\cite{colbert}, we also make the following optimizations to the encoder.
\begin{itemize}
\item Prepend the special token \texttt{[Q]} and \texttt{[D]} to the question tokens and sentence tokens respectively. This makes it easier for BERT encoder to identify the type of text.

\item We pad question tokens with BERT’s special \texttt{[mask]} token up to a fixed length. The token \texttt{[mask]} was used to mask the words to be predicted while pre-training. It allows BERT encoder to generate representations that match the given question's semantics by filling \texttt{[mask]} tokens to optimize the question representation.

\item Concatenate the question tokens after sentence tokens. It allows the question and sentences to interact with each other inside the BERT encoder.
\end{itemize}

\subsubsection{Relevance Scorer}
Using $\mathbf{H}^Q$ and $\mathbf{H}^D$, the retriever computes the relevance score of $D$ to $Q$ through a late interaction mechanism to allow token-level interaction by calculating token similarity as in ColBERT:
\begin{equation}
    \mathtt{sim}(\mathbf{H}^Q,\mathbf{H}^D)=\sum_{\mathbf{H}^Q_i \in \mathbf{H}^Q}\max_{\mathbf{H}^D_j \in \mathbf{H}^D}\frac{\mathbf{H}^Q_i \cdot{\mathbf{H}^D_j}^T}{\|\mathbf{H}^Q_i\|\|{\mathbf{H}^D_j}^T\|}
\end{equation}
where $\|\cdot\|$ represents vector modulus operator. We take top-$k$ sentences with the highest scores as the retrieval results.

Our retriever adopts RankNet pairwise loss~\cite{ranknet,optimizinghn}. Given the question $Q$, let $\mathbb{D}^+$ and $\mathbb{D}^-$ be the sets of all positive sentences and all negative sentences respectively. Our loss function is formulated as follows:
\begin{equation}
    \begin{aligned}
        L(Q,\mathbb{D}^+,\mathbb{D}^-)&=\\
    \sum_{D^+_i \in \mathbb{D}^+}&\sum_{D^-_j \in \mathbb{D}^-}\log(1+\frac{\exp(\mathtt{sim}(\mathbf{H}^Q,\mathbf{H}^{D^-_j}))}{\exp(\mathtt{sim}(\mathbf{H}^Q,{\mathbf{H}^{D^+_i}}))}).
    \end{aligned}
\end{equation}
With this loss function, we compare the scores of all positives and all negatives pairwisely to allow pairwise interaction between sentences. It aims to rank all positives ahead of all negatives in the retrieval results.

\subsection{Generator}\label{sec:generator}
Our generator extends FinQANet's generator by incorporating a dynamic reranker.

Given the question $Q$ and the retrieved sentences $\mathbb{D}^{\text{R}}=\{D^{\text{R}}_1,\cdots,D^{\text{R}}_k\}\subseteq\mathbb{D}'$, our generator aims to generate the arithmetic expression that the question implies. Figure~\ref{fig:generator_model}
 gives an overview of our generator~(top left and center).

\subsubsection{Encoder}
For question $Q=(u^Q_1,u^Q_2,\cdots,u^Q_{\lvert Q\rvert})$, we directly feed it into BERT to obtain its representation:
\begin{equation}
    \begin{aligned}
        \mathbf{H}^Q&=[\mathbf{h}^Q_1;\mathbf{h}^Q_2;\cdots;\mathbf{h}^Q_{\lvert Q\rvert+2}]\\
        &=\mathtt{BERT}(\texttt{[CLS]},u^Q_1,u^Q_2,\cdots,u^Q_{\lvert Q\rvert},\texttt{[SEP]})
    \end{aligned}
\end{equation}
and for each retrieved sentence $D^\text{R}_i=(u^{D^\text{R}_i}_1,u^{D^\text{R}_i}_2,\cdots,u^{D^\text{R}_i}_{\lvert D^\text{R}_i\rvert})$, in order to let it fully interact with $Q$ when encoding, we concatenate it with $Q$:
\begin{equation}
    \begin{aligned}
        \mathbf{H}^{D^\text{R}_i}&=[\mathbf{h}^{D^\text{R}_i}_1;\mathbf{h}^{D^\text{R}_i}_2;\cdots;\mathbf{h}^{D^\text{R}_i}_{\lvert {D^\text{R}_i}\rvert+\lvert Q\rvert+3}]\\
        &=\mathtt{BERT}(\texttt{[CLS]},u^{D^\text{R}_i}_1,u^{D^\text{R}_i}_2,\cdots,u^{D^\text{R}_i}_{\lvert D^\text{R}_i\rvert},\\
        &\indent\indent\indent\indent\texttt{[SEP]},u^Q_1,u^Q_2,\cdots,u^Q_{\lvert Q\rvert},\texttt{[SEP]}).
    \end{aligned}
    \label{eq:sentence_representation}
\end{equation}
We then concatenate the representation of the question and the representations of all the retrieved sentences:
\begin{equation}
    \begin{aligned}
        \mathbf{H}=\mathbf{H}^Q\Vert\mathbf{H}^{D^\text{R}_1}\Vert\cdots\Vert\mathbf{H}^{D^\text{R}_k}
    \end{aligned}
    \label{eq:input_representations}
\end{equation}
where $\Vert$ represents concatenation.

DSL contains predefined binary operators with numeric arguments~($\mathtt{ADD}$, $\mathtt{DIVIDE}$, etc.), operators with table row headers as arguments~($\mathtt{TABLE\_SUM}$, $\mathtt{TABLE\_MAX}$, etc.) which operate on a row, constant tokens ($\mathtt{CONST\_1}$, $\mathtt{CONST\_100}$, $\mathtt{CONST\_1000}$, etc.) and memory tokens representing the results of previous operations ($\mathtt{\#0}$, $\mathtt{\#1}$, etc.).
For all these tokens $U^{\text{DSL}}=\{u^{\text{DSL}}_1,u^{\text{DSL}}_2,\cdots,u^{\text{DSL}}_m\}$ in the DSL, we randomly initialise their representations $\mathbf{H}^{\text{DSL}}=[\mathbf{h}^{\text{DSL}}_1;\mathbf{h}^{\text{DSL}}_2;\cdots;\mathbf{h}^{\text{DSL}}_m]$ in training.

\subsubsection{Decoder}
We begin by introducing the attention mechanism, which will be used frequently in the following:
\begin{equation}
    \begin{aligned}
        \mathtt{Attention}(\mathbf{Q},\mathbf{K},\mathbf{V})=\mathtt{Softmax}(\mathbf{W}^Q\mathbf{Q}\mathbf{K}^T)\mathbf{V}
    \end{aligned}
\end{equation}
where $\mathbf{W}^Q$ is a matrix of learnable parameters.

At step $t$ of generation, the token representation sequence that has been generated is $\mathbf{Y}_t=[\mathbf{y}_1;\mathbf{y}_2;\cdots;\mathbf{y}_{t-1}]$
and the decoder output representation, denoted by $\hat{\mathbf{y}}_t$, is obtained by a LSTM network:
\begin{equation}
    \mathbf{\hat{y}}_{t}=\mathtt{LSTM}(\mathbf{y}_{t-1}).
\end{equation}
We randomly initialize $\mathbf{y}_0$ in training.
Our goal is to determine which token in the question, sentences, or DSL should be selected. In the following, unless otherwise stated, all variables are for step $t$ and we may omit the subscript $t$.

We calculate the attention representation of $\hat{\mathbf{y}}_t$ over $\mathbf{Y}_t$ and $\mathbf{H}^{\text{DSL}}\Vert\mathbf{H}$ to get a decoder history representation and a sequence representation, respectively:
\begin{equation}
    \begin{aligned}
        \mathbf{a}^\text{Y}=\mathtt{Attention}(\mathbf{Q}=\mathbf{\hat{y}}_t,\mathbf{K}&=\mathbf{Y}_t,\mathbf{V}=\mathbf{Y}_t)\\
    \mathbf{a}^\text{seq}=\mathtt{Attention}(\mathbf{Q}=\mathbf{\hat{y}}_t,\mathbf{K}&=\mathbf{H}^{\text{DSL}}\Vert\mathbf{H},\\
    \mathbf{V}&=\mathbf{H}^{\text{DSL}}\Vert\mathbf{H}).
    \end{aligned}
\end{equation}
Note that $\mathbf{\hat{y}}_t$, as the query of attention, is not included in $\mathbf{a}^\text{Y}$ and $\mathbf{a}^\text{seq}$, so we fuse them to get a context representation:
\begin{equation}
    \begin{aligned}
        \mathbf{h}^\text{c}=\mathtt{Layernorm}(\mathbf{W}^\text{c}(\mathbf{a}^\text{Y}\Vert\mathbf{a}^\text{seq}\Vert\mathbf{\hat{y}}_t))
    \end{aligned}
\end{equation}
where $\mathbf{W}^\text{c}$ is a matrix of learnable parameters.
For now, we have obtained the representations of all tokens in DSL, i.e., $\mathbf{H}^{\text{DSL}}$, and all tokens in $Q$ and $\mathbb{D}^\text{R}$, i.e., $\mathbf{H}$. We fuse them and compute the scores of all tokens by calculating the inner product of token representation and context representation:
\begin{equation}
    \begin{aligned}
        \mathbf{H}^\text{S}&=\mathbf{W}^\text{S}((\mathbf{H}^{\text{DSL}}\Vert\mathbf{H})\Vert((\mathbf{H}^{\text{DSL}}\Vert\mathbf{H})\circ \mathbf{a}^\text{seq}))\\
        \mathbf{s}^{\text{ctx}}&=\mathbf{H}^\text{S}\mathbf{h}^\text{c}
    \end{aligned}
    \label{eq:ctx_representations}
\end{equation}
in which $\circ$ is element-wise product and $\mathbf{W}^\text{S}$ is a matrix of learnable parameters. 
Note that in Equation~(\ref{eq:ctx_representations}) we duplicate the vector $\mathbf{a}^\text{seq}$ to form a matrix to be concatenated with other matrices, which for clarity is omitted in the equation.

We get the reranker scores $\mathbf{s}^{\text{r}}$ from our reranker which will be described in detail in Section \ref{sec:reranker}. 
The final scores of all tokens are represented by the element-wise product of context scores and reranker scores:
\begin{equation}
    \mathbf{s}=\mathbf{s}^{\text{ctx}}\circ\mathbf{s}^{\text{r}}.
    \label{eq:scores}
\end{equation}

We optimize the cross-entropy loss during training:
\begin{equation}
    \begin{aligned}
        \mathcal{L}=-\log\frac{\exp(s^\text{gold})}{\sum_{tok\in U^{\text{DSL}}\cup U^{\text{num}}}\exp(s_{tok})}
    \end{aligned}
\end{equation}
where $U^{\text{num}}$ is the set of all numeric tokens, $s_{tok}$ is the final score of a token computed by Equation~(\ref{eq:scores}), and $s^\text{gold}$ is the final score of the ground truth token.

We finally select the predicted token by $\mathbf{s}$:
\begin{equation}
    \mathbf{y}_t=\mathbf{H}^\text{S}_{\arg \max \mathbf{s}}
\end{equation}
i.e., the representation of the ($\arg \max \mathbf{s}$)-th token in $\mathbf{H}^\text{S}$. We use teacher forcing mechanism during training, i.e., we use the representation of the ground truth token as $\mathbf{y}_t$.

\subsection{Reranker}\label{sec:reranker}
In this module, we aim to find the sentence that is of interest to the generator at step $t$, i.e., the one in which the next numeric argument of the arithmetic expression is located.

\subsubsection{Reranker Space}
The reranker space, i.e., the objects that need to be reranked, consists of representations of the question and retrieved sentences.

We construct the reranker space as follows.
We use dual attention~\cite{dualattention} to enhance the interaction between the question representation $\mathbf{H}^Q$ and each sentence representation $\mathbf{H}^{D_i^\text{R}}$ computed by Equation~(\ref{eq:sentence_representation}):
\begin{equation}
    \begin{aligned}
        \hat{\mathbf{H}}^{Q_i},\hat{\mathbf{H}}^{D_i^\text{R}}=\mathtt{DualAttention}(\mathbf{H}^Q&,\mathbf{H}^{D_i^\text{R}}).
    \end{aligned}
\end{equation}
$\hat{\mathbf{H}}^{Q_i}$ is the token sequence representations of $Q$ after interacting with $D_i^\text{R}\in\mathbb{D}^{\text{R}}$.
Then mean-pooling in token dimension is applied to obtain the representations of $Q$ and $D_i^\text{R}$:
\begin{equation}
    \begin{aligned}
        \mathbf{h}^{Q_i}&=\mathtt{MeanPooling}(\hat{\mathbf{H}}^{Q_i})\\
        \mathbf{h}^{D_i^\text{R}}&=\mathtt{MeanPooling}(\hat{\mathbf{H}}^{D_i^\text{R}})\\
    \end{aligned}
    \label{eq:sentence_vector_representation}
\end{equation}
We combine the $k$ representations of $Q$ over $k$~sentences to obtain a representation of $Q$ by attention mechanism:
\begin{equation}
    \begin{aligned}
        \mathbf{h}^Q=\mathtt{Attention}(\mathbf{Q}=\hat{\mathbf{y}}_t,\mathbf{K}&=[\mathbf{h}^{Q_1};\mathbf{h}^{Q_2};\cdots,\mathbf{h}^{Q_k}],\\
        \mathbf{V}&=[\mathbf{h}^{Q_1};\mathbf{h}^{Q_2};\cdots,\mathbf{h}^{Q_k}]).
    \end{aligned}
    \label{eq:question_representation}
\end{equation}
The reranker space contains $\mathbf{h}^Q, \mathbf{h}^{D_1^\text{R}}, \ldots, \mathbf{h}^{D_k^\text{R}}$.

\subsubsection{Reranker Query}
The query used to rerank is dynamically generated, not only in relation to the expression that has been generated, i.e., $\mathbf{Y}_t$, but also in relation to the source sentences from which the tokens in the expression are extracted.
We first locate the source sentence of each token in the generated expression. We take the representation of its source sentence if the generated token is numeric, or keep the representation of the token if it is from the DSL, formally:
\begin{equation} 
    \mathtt{source}(\mathbf{y}_i)=\\ \left\{
    \begin{aligned}
        \mathbf{y}_i,\indent &\text{if }\mathbf{y}_i\in\mathbf{H}^{\text{DSL}},\\
        \mathbf{h}^{D},\indent &\text{if }\mathbf{y}_i \text{ comes from } D\in\mathbb{D}^{\text{QR}}
    \end{aligned}
    \right.
\end{equation}
where $\mathbb{D}^{\text{QR}}=\{Q\}\cup\mathbb{D}^{\text{R}}$ and $\mathbf{h}^{D}$ refers to $\mathbf{h}^{Q}$ computed by Equation (\ref{eq:question_representation}) or $\mathbf{h}^{D^{\text{R}}_i}$ computed by Equation (\ref{eq:sentence_vector_representation}).
We obtain the source sentence representations of $\mathbf{Y}_t$ by
\begin{equation}
    \begin{aligned}
        \mathbf{H}^{\text{src}}_t=\mathbf{H}^{\text{src}}_{t-1};\mathtt{source}(\mathbf{y}_{t-1})
    \end{aligned}
\end{equation}
which refers to appending the source sentence representation of $\mathbf{y}_{t-1}$ to the source sentence representations of $\mathbf{Y}_{t-1}$ which were calculated at the previous step.

We use the source sentence representations of generated tokens to construct a query to rerank all the sentences at the current step.
We compute the attention representation of $\hat{\mathbf{y}}_t$ over $\mathbf{H}^{\text{src}}_t$ to obtain the reranker query representation $\mathbf{Q}^\text{s}$:
\begin{equation}
    \begin{aligned}
        \mathbf{Q}^\text{s}=\mathtt{Attention}(\mathbf{Q}=\hat{\mathbf{y}}_t,\mathbf{K}=\mathbf{H}^{\text{src}}_t,\mathbf{V}=\mathbf{H}^{\text{src}}_t).
    \end{aligned}
\end{equation}

\subsubsection{Reranker Scorer}
With the reranker query $\mathbf{Q}^\text{s}$ and the reranker space $\{\mathbf{h}^Q,\mathbf{h}^{D_1},\cdots,\mathbf{h}^{D_k}\}$, we re-compute the relevance score of each sentence by:
\begin{equation}
    \begin{aligned}
        s^Q&=\mathtt{Linear}(\mathtt{Tanh}(\mathtt{Linear}([\mathbf{Q}^\text{s}\Vert\mathbf{h}^{Q}])))\\
        s^{D_i^\text{R}}&=\mathtt{Linear}(\mathtt{Tanh}(\mathtt{Linear}([\mathbf{Q}^\text{s}\Vert\mathbf{h}^{D_i^\text{R}}]))), i\in \{1,\cdots,k\}
    \end{aligned}
\end{equation}
The reranker scores in Equation~(\ref{eq:scores}) are given by
\begin{equation}
    \mathbf{s}^{\text{r}}=\mathtt{Expand}(\mathtt{Softmax}([s^Q,s^{D_1^\text{R}},\cdots,s^{D_k^\text{R}}])).
\end{equation}
Note that $s^Q,s^{D_1^\text{R}},\cdots,s^{D_k^\text{R}}$ are for scoring sentences while we need to obtain reranker scores for tokens, so an $\mathtt{Expand}$ operation is applied which assigns the score of each sentence to every token in that sentence. The scores of tokens in the DSL are fixed to 1.0.
\section{Experiments}\label{sec:experiments}

\subsection{Datasets}
To the best of our knowledge, there were two datasets requiring generating arithmetic expressions for numerical reasoning over tabular and textual data: FinQA~\cite{finqa} and MultiHiertt~\cite{multihiertt}. We used FinQA in our experiments. We did not use MultiHiertt because it was new and yet to be further polished after contacting its authors.

FinQA contains 8,281 financial questions which were collected from financial reports of S\&P 500 companies. Questions were divided into 6,251~(75\%) for training, 883~(10\%) for development, and 1,147~(15\%) for testing. Each question contains a table where the average number of rows is 6.36. The average number of sentences accompanying a question is 24.32. The total number of tokens in the table and sentences accompanying a question averages 687.53, with a maximum value of 2,679.


\subsection{Implementation Details}
We experimented on NVIDIA V100 (32GB) GPUs and Ubuntu 18.04. Our implementations were based on PyTorch~1.11. We selected models on the development set.

For the encoder in our retriever, we used BERT-base-uncased with $\text{hidden layer}=12$, $\text{hidden units}=768$, and $\text{attention heads}=12$. We set $\text{epoch}=8$ and $\text{seed}=8$. We used the Adam optimizer with $\text{learning rate}=2e-5$ selected from $\{1.5e-5,2e-5\}$ with $\text{batch size}=16$. We set $k=3$ and $\text{max sequence length}=256$.

For the encoder in our generator, we used two pretrained models: BERT-base-uncased and RoBERTa-large. RoBERTa-large is with $\text{hidden layer}=24$, $\text{hidden units}=1\text{,}024$, and $\text{attention heads}=16$. 
In our reranker and generator, we set $\text{epoch}=300$ and $\text{seed}=8$. We used the Adam optimizer with $\text{learning rate}=2e-5$ for BERT-base-uncased and $\text{learning rate}=1.5e-5$ for RoBERTa-large, both selected from $\{7e-6,1.5e-5,2e-5\}$. We set $\text{batch size}=16$ for BERT-base-uncased and $\text{batch size}=24$ for RoBERTa-large, both selected from $\{16,24,32\}$. We set $\text{max sequence length}=256$.

\subsection{Baselines}
We compared our proposed model DyRRen with known methods and results reported in the literature.

We mainly compared DyRRen with FinQANet, which is a state-of-the-art retriever-generator model specifically designed for numerical reasoning over tabular and textual data. We compared with FinQANet's versions on FinBERT~\cite{finbert}, BERT-base-uncased, and RoBERTa-large. \citet{finqa} provided the results of all the three versions on FinQA. 

Longformer~\cite{longformer} is specifically designed to handle long documents, and was used in our experiments to verify the necessity to include a retriever in our framework. NeRd~\cite{nerd} is an expression generator based on a pointer network and has achieved competitive results on MathQA~\cite{mathqa} and DROP~\cite{drop}. We used the results of Longformer and NeRd reported in~\citet{finqa}.

\subsection{Evaluation Metrics}
Following the literature, we measured program accuracy~(PA), i.e., proportion of correctly generated arithmetic expressions, and execution accuracy~(EA), i.e., proportion of correct final answers. Note that sometimes a generated expression is different from the corresponding ground truth but their answers are equivalent, so execution accuracy is always at least as high as program accuracy.

\subsection{Main Results}
\begin{table}[t]
    \centering
    \small
    \begin{tabular}{|l|cccc|}
        \hline
         & \multicolumn{2}{c}{Dev}  & \multicolumn{2}{c|}{Test} \\
        Methods & EA & PA & EA & PA \\
        \hline
        Longformer$_\texttt{base}$ & 23.83 & 22.56 & 21.90 & 20.48 \\
        NeRd & 47.53 & 45.37 & 48.57 & 46.76 \\
        FinQANet$_\texttt{FinBERT}$ & 46.64 & 44.11 & 50.10 & 47.52 \\
        FinQANet$_\texttt{BERT}$ & 49.91 & 47.15 & 50.00 & 48.00 \\
        FinQANet$_\texttt{RoBERTa}$ & 61.22 & 58.05 & 61.24 & 58.86 \\
        \hline
        DyRRen$_\texttt{BERT}$ & 61.16 & 58.32 & 59.37 & 57.54 \\
        DyRRen$_\texttt{RoBERTa}$ & \textbf{66.82} & \textbf{63.87} & \textbf{63.30} & \textbf{61.29} \\
        \hline 
        Human Expert & -- & -- & 91.16 & 87.49 \\
        General Crowd & -- & -- & 50.68 & 48.17 \\
        \hline
    \end{tabular}
    \caption{Comparison with baselines on FinQA. $\texttt{BERT}$ refers to BERT-base-uncased, $\texttt{RoBERTa}$ refers to RoBERTa-large and $\texttt{FinBERT}$ is a BERT-like model pretrained in the financial domain.}
    \label{table:main_results_finqa}
\end{table}

On the test set, as shown in Table~\ref{table:main_results_finqa}, DyRRen outperformed all the baselines by at least 9.37\% of EA and 9.54\% of PA with BERT and at least 2.06\% of EA and 2.43\% of PA with RoBERTa. The differences were statistically significant under $p<0.01$. Besides, both DyRRen$_\texttt{BERT}$ and DyRRen$_\texttt{RoBERTa}$ exceeded general crowd~(50.68\% of EA and 48.17\% of PA), although they were still not comparable with human expert~(91.16\% of EA and 87.49\% of PA).


It is notable that DyRRen$_\texttt{BERT}$ achieved similar performance to FinQANet$_\texttt{RoBERTa}$, while the number of parameters in FinQANet$_\texttt{RoBERTa}$~($\sim$380M) is even 2.7 times that of DyRRen$_\texttt{BERT}$~($\sim$139M).

\subsection{Case Study}
\begin{figure}[t]
    \centering
    \includegraphics[width=0.9\linewidth]{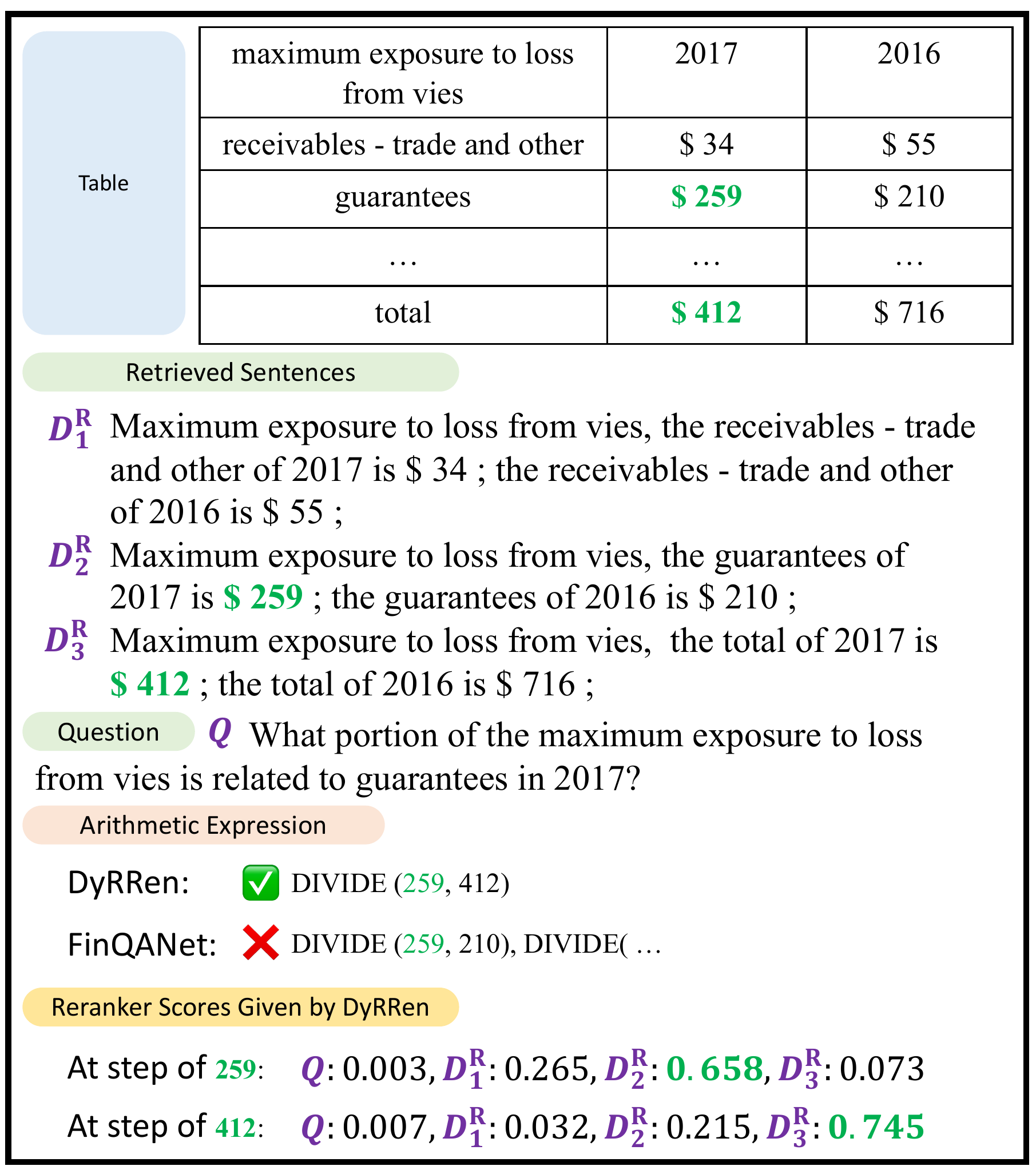}
    \caption{Case study sampled from FinQA.}
    \label{fig:case_study}
\end{figure}
As shown in Figure~\ref{fig:case_study}, we sampled a question from FinQA for which the retrievers of both DyRRen and FinQANet~(BERT versions) successfully retrieved the sentences relevant to the question, i.e., $D_2^\text{R}$ and $D_3^\text{R}$. It can be seen that at the beginning, both DyRRen and FinQANet succeeded in generating the correct operator \texttt{DIVIDE} and extracting the first argument ``259" from $D^{\text{R}}_2$, and our reranker in DyRRen exhibited the highest confidence to $D^{\text{R}}_2$~(0.658). When generating the second argument, our reranker assigned a higher score~(0.745) to $D^{\text{R}}_3$ containing the correct argument ``412", successfully switching the generator's attention to the correct sentence, while FinQANet still stayed in $D^{\text{R}}_2$ and selects ``210", thus failing to answer this question.

\subsection{Ablation Study}
We compared DyRRen with two variants:
\begin{itemize}
    \item DyRRen$_\texttt{no-reranker}$, which removes the reranker module from our retrieval-ranker-generator framework.
    \item DyRRen$_\texttt{vanilla-retriever}$, which replaces our retriever with FinQANet's retriever to verify the effectiveness of the optimizations we have made to our retriever.
\end{itemize}

On FinQA, as shown in Table~\ref{table:ablation_results_finqa}, DyRRen$_\texttt{no-reranker}$ was 1.48\% lower in EA and 2.00\% lower in PA than DyRRen on the basis of BERT on the test set, demonstrating the usefulness of our reranker. The performance of DyRRen$_\texttt{no-reranker}$~(RoBERTa version) also noticeably dropped on the dev set~(by 0.98\% in EA and 1.02\% in PA), while the differences on the test set were neglectable. We thought it was the more powerful RoBERTa which counterbalanced some gain from our reranker.

The performance of DyRRen$_\texttt{vanilla-retriever}$ decreased on both BERT and RoBERTa based experiments. For example, DyRRen$_\texttt{vanilla-retriever}$ (BERT) was 3.22\% lower in EA and 3.75\% lower in PA than DyRRen on the test set, demonstrating the effectiveness of our retriever.

We also tried to incorporate models pre-trained on tables such as TaPas~\cite{tapas} into our encoder, but did not obtain better results.


\begin{table}[t]
    \centering
    \small
    \begin{tabular}{|l|cccc|}
        \hline
         & \multicolumn{2}{c}{Dev}  & \multicolumn{2}{c|}{Test} \\
        Methods & EA & PA & EA & PA \\
        \hline
        \textbf{\texttt{BERT}} & & & & \\
        DyRRen & 61.16 & 58.32 & 59.37 & 57.54 \\
        DyRRen$_\texttt{no-reranker}$ & 60.82 & 58.66 & 57.89 & 55.54 \\
        DyRRen$_\texttt{vanilla-retriever}$ & 58.78 & 56.17 & 56.15 & 53.79 \\
        \textbf{\texttt{RoBERTa}} & & & & \\
        DyRRen & 66.82 & 63.87 & 63.30 & 61.29 \\
        DyRRen$_\texttt{no-reranker}$ & 65.80 & 62.85 & 62.86 & 61.46 \\
        DyRRen$_\texttt{vanilla-retriever}$ & 62.85 & 60.48 & 60.33 & 57.89 \\
        \hline
    \end{tabular}
    \caption{Ablation study on FinQA.}
    \label{table:ablation_results_finqa}
\end{table}

\subsection{Fine-Grained Results}
We conducted fine-grained comparison between DyRRen$_\texttt{BERT}$ and FinQANet$_\texttt{BERT}$, as shown in Table~\ref{table:performance_analysis_finqa}.
\subsubsection{Questions categorized by question type}
Our model made notable improvements over FinQANet on all the three types of questions. We also observed that both models performed the best on table-only questions, i.e., those that can be answered only based on tables. This may be due to the fact that row sentences have good structure as they are converted by using templates, so they are relatively easy to be understood. Both models performed the worst on table-sentence questions, indicating that there is still room for improving hybrid data understanding.

\subsubsection{Questions categorized by expression length}
Our reranker helped our model focus on the unique supporting sentence for the current step rather than on all the retrieved sentences, and it was consistently superior on questions of all expression lengths. We also found that both models performed poorly when expression length exceeded two. These ``long-distance reasoning" questions are very challenging for existing methods and await further study.
\begin{table}[th]
    \centering
    \small
    \begin{tabular}{|l|cccc|}
        \hline
         & \multicolumn{2}{c}{Dev}  & \multicolumn{2}{c|}{Test} \\
         & EA & PA & EA & PA \\
        \hline
        \multicolumn{5}{|l|}{DyRRen$_\texttt{BERT}$} \\
        \hline
        \textbf{\texttt{Question type}} & & & & \\
        table-only & 72.51 & 69.37 & 68.98 & 66.71 \\
        sentence-only & 46.46 & 44.44 & 49.47 & 48.76 \\
        table-sentence & 38.46 & 35.66 & 34.18 & 32.28 \\
        \textbf{\texttt{Expression length}}  & & & & \\
        $1$ & 66.16 & 63.67 & 64.37 & 63.00 \\
        $2$ & 60.98 & 57.49 & 57.46 & 54.77 \\
        $>2$ & 26.03 & 23.29 & 29.76 & 28.57 \\
        \hline
        \multicolumn{5}{|l|}{FinQANet$_\texttt{BERT}$} \\
        \hline
        \textbf{\texttt{Question type}} & & & & \\
        table-only & 61.62 & 58.86 & 58.22 & 55.67 \\
        sentence-only & 37.37 & 35.86 & 40.64 & 39.93 \\
        table-sentence & 28.67 & 27.27 & 31.01 & 30.38 \\
        \textbf{\texttt{Expression length}}  & & & & \\
        $1$ & 52.77 & 51.05 & 52.14 & 50.76 \\
        $2$ & 54.01 & 50.52 & 53.55 & 51.10 \\
        $>2$ & 24.66 & 23.29 & 17.86 & 15.48 \\
        \hline
    \end{tabular}
    \caption{Fine-grained comparison with FinQANet on FinQA by question type and expression length.}
    \label{table:performance_analysis_finqa}
\end{table}

\subsection{Error Analysis}
We randomly sampled 50 questions for which our model failed to generate the correct arithmetic expression. The errors caused by our retriever constituted about 20\% of all errors, e.g., the retriever failed to retrieve all the supporting sentences. About 14\% of the errors were caused by the reranker misleading the generator, i.e., the reranker assigned improper scores to sentences in some generation step. The rest of the errors~(66\%) were due to the generator. There were many reasons for these generation errors, 
one of the main ones being that the generator failed to actively switch attention between sentences, and the failure was too strong to be counterbalanced by our reranker.
Other reasons include incorrect prediction of operators, or the generator became lost on questions that require multi-step numerical reasoning or require external financial knowledge.

\section{Related Work}\label{sec:related_work}

\subsection{Question Answering over Hybrid Data}
QA that requires reasoning over a hybrid context, i.e., tabular and textual data, is a task that has emerged in the last two years. Two of the first datasets focusing on this task are HybridQA~\cite{hybridqa} and TSQA~\cite{tsqa}. The former is an extractive QA dataset focusing more on multi-hop reasoning while the latter is a multiple-choice QA dataset more focusing on numerical reasoning. \citet{mate} proposes a Transformer-based model with row-wise and column-wise attention and \cite{kumar2021multi} applies multiple-instance training to solve HybridQA. \citet{tsqa} designs a number of templates containing numerical operations to convert tables to sentences, and then uses a knowledge-injected pretrained language model to solve TSQA.

Following these efforts, a growing number of more challenging hybrid QA datasets have emerged. OTT-QA~\cite{ottqa} and NQ-Tables~\cite{nqtables} place hybrid QA in an open-domain setting where tables are not given but need to be retrieved from a table corpus. \citet{nqtables} solves this problem by dense retrieval.
Later, it was found that tables in the financial domain carry a lot of numerical content and are suitable for QA tasks that require numerical reasoning over tabular and textual data. TAT-QA~\cite{tat-qa}, FinQA~\cite{finqa}, and MultiHiertt~\cite{multihiertt} are such datasets containing QA pairs from financial reports. TAT-QA requires the answer to a question to be extracted or calculated from given sentences and tables. FinQA and MultiHiertt require the generation of an arithmetic expression, making the question answering process more interpretable.

So far, FinQANet~\cite{finqa}, which adopts a retrieval-generator architecture, remains state of the art for solving hybrid QA that requires the generation of arithmetic expressions. The major architectural difference between FinQANet and our DyRRen is that we fuse the idea of reranking into each generation step, i.e., our dynamic reranker, to improve the effectiveness of our generator.

\subsection{Numerical Reasoning in QA}
While current pretrained language models~\cite{lin2020birds} could not perform well on numerical reasoning, there is a growing interest in numerical reasoning based QA. DROP~\cite{drop} requires mathematical operations such as adding, subtracting, counting, and sorting. Later work shows that it can be solved by semantic parsing~\cite{gupta2019neural}, graph-base method~\cite{chen2020question}, neural module network~\cite{zhou2022opera}, or enhancing number-related context understanding~\cite{kim2022exploiting}.

Math word problem~(MWP)~\cite{hosseini2014learning,huang2016well,mathqa} is a fundamental and challenging task which has been introduced for years. The task is to, given a short textual question, calculate the answer by generating an arithmetic expression. There are two major differences between MWP and the task considered in this paper: 1) MWP contains only a short textual question without tables and supporting sentences, so solving MWPs does not require to handle tabular data nor the retrieval problem. 2) Questions in MWPs are described in a controlled manner, exhibiting strong regularity so some of the early work could use statistical learning~\cite{hosseini2014learning,kushman2014learning} or rule/template-based methods~\cite{shi2015automatically,wang2019template} to achieve fairly good results. \citet{huang2021recall} proposes a recall learning framework for solving MWPs, a framework that recalls problems with similar expression structures, benefiting from the regularity of MWPs. There are also graph/tree based~\cite{wu2021edge,jie2022learning} or multiple-task learning~\cite{qin2021neural,shen2021generate} methods. However, due to the above-mentioned differences between two tasks, these solutions to MWPs are unlikely to obtain promising results on our task.
\section{Conclusion}\label{sec:conclusion}
We present DyRRen, a dynamic retriever-reranker-generator model for numerical reasoning over tabular and textual data. We believe that the idea of incorporating a reranker to enhance generation has the potential to be applicable to other generation tasks where the input contains multiple sentences, including machine translation, machine writing, summarization, etc., which we will explore as future work.

Our error analysis also reveals some limitations of our model. Our reranker is able to help correct the selection of numerical arguments of operators, but is unable to handle incorrect prediction of operators. 
We will extend our reranker to handle operators in future work.
We are also ready to experiment on more datasets to evaluate the generalizability of our model, for instance, when the emerging MultiHiertt dataset becomes stable.

\section*{Acknowledgments}
This work was supported in part by the NSFC (62072224) and in part by the CCF-Baidu Open Fund.

\bibliography{aaai23.bib}

\begin{thebibliography}{37}
\providecommand{\natexlab}[1]{#1}

\bibitem[{Amini et~al.(2019)Amini, Gabriel, Lin, Koncel{-}Kedziorski, Choi, and
  Hajishirzi}]{mathqa}
Amini, A.; Gabriel, S.; Lin, S.; Koncel{-}Kedziorski, R.; Choi, Y.; and
  Hajishirzi, H. 2019.
\newblock MathQA: Towards Interpretable Math Word Problem Solving with
  Operation-Based Formalisms.
\newblock In \emph{Proceedings of the 2019 Conference of the North American
  Chapter of the Association for Computational Linguistics: Human Language
  Technologies, {NAACL-HLT} 2019, Minneapolis, MN, USA, June 2-7, 2019, Volume
  1 (Long and Short Papers)}, 2357--2367.

\bibitem[{Araci(2019)}]{finbert}
Araci, D. 2019.
\newblock FinBERT: Financial Sentiment Analysis with Pre-trained Language
  Models.
\newblock \emph{CoRR}, abs/1908.10063.

\bibitem[{Beltagy, Peters, and Cohan(2020)}]{longformer}
Beltagy, I.; Peters, M.~E.; and Cohan, A. 2020.
\newblock Longformer: The Long-Document Transformer.
\newblock \emph{CoRR}, abs/2004.05150.

\bibitem[{Burges(2010)}]{ranknet}
Burges, C.~J. 2010.
\newblock From RankNet to LambdaRank to LambdaMART: An Overview.
\newblock Technical Report MSR-TR-2010-82, Microsoft Research.

\bibitem[{Chen et~al.(2020{\natexlab{a}})Chen, Xu, Cheng, Xiaochuan, Zhang,
  Song, Wang, Qi, and Chu}]{chen2020question}
Chen, K.; Xu, W.; Cheng, X.; Xiaochuan, Z.; Zhang, Y.; Song, L.; Wang, T.; Qi,
  Y.; and Chu, W. 2020{\natexlab{a}}.
\newblock Question Directed Graph Attention Network for Numerical Reasoning
  over Text.
\newblock In \emph{Proceedings of the 2020 Conference on Empirical Methods in
  Natural Language Processing, {EMNLP} 2020, Online, November 16-20, 2020},
  6759--6768.

\bibitem[{Chen et~al.(2021{\natexlab{a}})Chen, Chang, Schlinger, Wang, and
  Cohen}]{ottqa}
Chen, W.; Chang, M.; Schlinger, E.; Wang, W.~Y.; and Cohen, W.~W.
  2021{\natexlab{a}}.
\newblock Open Question Answering over Tables and Text.
\newblock In \emph{9th International Conference on Learning Representations,
  {ICLR} 2021, Virtual Event, Austria, May 3-7, 2021}.

\bibitem[{Chen et~al.(2020{\natexlab{b}})Chen, Zha, Chen, Xiong, Wang, and
  Wang}]{hybridqa}
Chen, W.; Zha, H.; Chen, Z.; Xiong, W.; Wang, H.; and Wang, W.~Y.
  2020{\natexlab{b}}.
\newblock HybridQA: {A} Dataset of Multi-Hop Question Answering over Tabular
  and Textual Data.
\newblock In \emph{Findings of the Association for Computational Linguistics:
  {EMNLP} 2020, Online Event, 16-20 November 2020}, volume {EMNLP} 2020,
  1026--1036.

\bibitem[{Chen et~al.(2020{\natexlab{c}})Chen, Liang, Yu, Zhou, Song, and
  Le}]{nerd}
Chen, X.; Liang, C.; Yu, A.~W.; Zhou, D.; Song, D.; and Le, Q.~V.
  2020{\natexlab{c}}.
\newblock Neural Symbolic Reader: Scalable Integration of Distributed and
  Symbolic Representations for Reading Comprehension.
\newblock In \emph{8th International Conference on Learning Representations,
  {ICLR} 2020, Addis Ababa, Ethiopia, April 26-30, 2020}.

\bibitem[{Chen et~al.(2021{\natexlab{b}})Chen, Chen, Smiley, Shah, Borova,
  Langdon, Moussa, Beane, Huang, Routledge, and Wang}]{finqa}
Chen, Z.; Chen, W.; Smiley, C.; Shah, S.; Borova, I.; Langdon, D.; Moussa, R.;
  Beane, M.; Huang, T.; Routledge, B.~R.; and Wang, W.~Y. 2021{\natexlab{b}}.
\newblock FinQA: {A} Dataset of Numerical Reasoning over Financial Data.
\newblock In \emph{Proceedings of the 2021 Conference on Empirical Methods in
  Natural Language Processing, {EMNLP} 2021, Virtual Event / Punta Cana,
  Dominican Republic, 7-11 November, 2021}, 3697--3711.

\bibitem[{Devlin et~al.(2019)Devlin, Chang, Lee, and Toutanova}]{bert}
Devlin, J.; Chang, M.; Lee, K.; and Toutanova, K. 2019.
\newblock {BERT:} Pre-training of Deep Bidirectional Transformers for Language
  Understanding.
\newblock In \emph{Proceedings of the 2019 Conference of the North American
  Chapter of the Association for Computational Linguistics: Human Language
  Technologies, {NAACL-HLT} 2019, Minneapolis, MN, USA, June 2-7, 2019, Volume
  1 (Long and Short Papers)}, 4171--4186.

\bibitem[{Dua et~al.(2019)Dua, Wang, Dasigi, Stanovsky, Singh, and
  Gardner}]{drop}
Dua, D.; Wang, Y.; Dasigi, P.; Stanovsky, G.; Singh, S.; and Gardner, M. 2019.
\newblock {DROP:} {A} Reading Comprehension Benchmark Requiring Discrete
  Reasoning Over Paragraphs.
\newblock In \emph{Proceedings of the 2019 Conference of the North American
  Chapter of the Association for Computational Linguistics: Human Language
  Technologies, {NAACL-HLT} 2019, Minneapolis, MN, USA, June 2-7, 2019, Volume
  1 (Long and Short Papers)}, 2368--2378.

\bibitem[{Eisenschlos et~al.(2021)Eisenschlos, Gor, M{\"{u}}ller, and
  Cohen}]{mate}
Eisenschlos, J.; Gor, M.; M{\"{u}}ller, T.; and Cohen, W.~W. 2021.
\newblock {MATE:} Multi-view Attention for Table Transformer Efficiency.
\newblock In \emph{Proceedings of the 2021 Conference on Empirical Methods in
  Natural Language Processing, {EMNLP} 2021, Virtual Event / Punta Cana,
  Dominican Republic, 7-11 November, 2021}, 7606--7619.

\bibitem[{Gupta et~al.(2020)Gupta, Lin, Roth, Singh, and
  Gardner}]{gupta2019neural}
Gupta, N.; Lin, K.; Roth, D.; Singh, S.; and Gardner, M. 2020.
\newblock Neural Module Networks for Reasoning over Text.
\newblock In \emph{8th International Conference on Learning Representations,
  {ICLR} 2020, Addis Ababa, Ethiopia, April 26-30, 2020}.

\bibitem[{Herzig et~al.(2021)Herzig, M{\"{u}}ller, Krichene, and
  Eisenschlos}]{nqtables}
Herzig, J.; M{\"{u}}ller, T.; Krichene, S.; and Eisenschlos, J. 2021.
\newblock Open Domain Question Answering over Tables via Dense Retrieval.
\newblock In \emph{Proceedings of the 2021 Conference of the North American
  Chapter of the Association for Computational Linguistics: Human Language
  Technologies, {NAACL-HLT} 2021, Online, June 6-11, 2021}, 512--519.

\bibitem[{Herzig et~al.(2020)Herzig, Nowak, M{\"{u}}ller, Piccinno, and
  Eisenschlos}]{tapas}
Herzig, J.; Nowak, P.~K.; M{\"{u}}ller, T.; Piccinno, F.; and Eisenschlos,
  J.~M. 2020.
\newblock TaPas: Weakly Supervised Table Parsing via Pre-training.
\newblock In \emph{Proceedings of the 58th Annual Meeting of the Association
  for Computational Linguistics, {ACL} 2020, Online, July 5-10, 2020},
  4320--4333.

\bibitem[{Hochreiter and Schmidhuber(1997)}]{lstm}
Hochreiter, S.; and Schmidhuber, J. 1997.
\newblock Long Short-Term Memory.
\newblock \emph{Neural Comput.}, 9(8): 1735--1780.

\bibitem[{Hosseini et~al.(2014)Hosseini, Hajishirzi, Etzioni, and
  Kushman}]{hosseini2014learning}
Hosseini, M.~J.; Hajishirzi, H.; Etzioni, O.; and Kushman, N. 2014.
\newblock Learning to Solve Arithmetic Word Problems with Verb Categorization.
\newblock In \emph{Proceedings of the 2014 Conference on Empirical Methods in
  Natural Language Processing, {EMNLP} 2014, October 25-29, 2014, Doha, Qatar,
  {A} meeting of SIGDAT, a Special Interest Group of the {ACL}}, 523--533.

\bibitem[{Huang et~al.(2016)Huang, Shi, Lin, Yin, and Ma}]{huang2016well}
Huang, D.; Shi, S.; Lin, C.; Yin, J.; and Ma, W. 2016.
\newblock How well do Computers Solve Math Word Problems? Large-Scale Dataset
  Construction and Evaluation.
\newblock In \emph{Proceedings of the 54th Annual Meeting of the Association
  for Computational Linguistics, {ACL} 2016, August 7-12, 2016, Berlin,
  Germany, Volume 1: Long Papers}.

\bibitem[{Huang et~al.(2021)Huang, Wang, Xu, Cao, and Yang}]{huang2021recall}
Huang, S.; Wang, J.; Xu, J.; Cao, D.; and Yang, M. 2021.
\newblock Recall and Learn: {A} Memory-augmented Solver for Math Word Problems.
\newblock In \emph{Findings of the Association for Computational Linguistics:
  {EMNLP} 2021, Virtual Event / Punta Cana, Dominican Republic, 16-20 November,
  2021}, 786--796.

\bibitem[{Jie, Li, and Lu(2022)}]{jie2022learning}
Jie, Z.; Li, J.; and Lu, W. 2022.
\newblock Learning to Reason Deductively: Math Word Problem Solving as Complex
  Relation Extraction.
\newblock In \emph{Proceedings of the 60th Annual Meeting of the Association
  for Computational Linguistics (Volume 1: Long Papers), {ACL} 2022, Dublin,
  Ireland, May 22-27, 2022}, 5944--5955.

\bibitem[{Khattab and Zaharia(2020)}]{colbert}
Khattab, O.; and Zaharia, M. 2020.
\newblock ColBERT: Efficient and Effective Passage Search via Contextualized
  Late Interaction over {BERT}.
\newblock In \emph{Proceedings of the 43rd International {ACM} {SIGIR}
  conference on research and development in Information Retrieval, {SIGIR}
  2020, Virtual Event, China, July 25-30, 2020}, 39--48.

\bibitem[{Kim et~al.(2022)Kim, Kang, Kim, Hong, and Myaeng}]{kim2022exploiting}
Kim, J.; Kang, J.; Kim, K.; Hong, G.; and Myaeng, S. 2022.
\newblock Exploiting Numerical-Contextual Knowledge to Improve Numerical
  Reasoning in Question Answering.
\newblock In \emph{Findings of the Association for Computational Linguistics:
  {NAACL} 2022, Seattle, WA, United States, July 10-15, 2022}, 1811--1821.

\bibitem[{Kumar et~al.(2021)Kumar, Chemmengath, Gupta, Sen, Bharadwaj, and
  Chakrabarti}]{kumar2021multi}
Kumar, V.; Chemmengath, S.~A.; Gupta, Y.; Sen, J.; Bharadwaj, S.; and
  Chakrabarti, S. 2021.
\newblock Multi-Instance Training for Question Answering Across Table and
  Linked Text.
\newblock \emph{CoRR}, abs/2112.07337.

\bibitem[{Kushman et~al.(2014)Kushman, Zettlemoyer, Barzilay, and
  Artzi}]{kushman2014learning}
Kushman, N.; Zettlemoyer, L.; Barzilay, R.; and Artzi, Y. 2014.
\newblock Learning to Automatically Solve Algebra Word Problems.
\newblock In \emph{Proceedings of the 52nd Annual Meeting of the Association
  for Computational Linguistics, {ACL} 2014, June 22-27, 2014, Baltimore, MD,
  USA, Volume 1: Long Papers}, 271--281.

\bibitem[{Li, Sun, and Cheng(2021)}]{tsqa}
Li, X.; Sun, Y.; and Cheng, G. 2021.
\newblock {TSQA:} Tabular Scenario Based Question Answering.
\newblock In \emph{Thirty-Fifth {AAAI} Conference on Artificial Intelligence,
  {AAAI} 2021, Thirty-Third Conference on Innovative Applications of Artificial
  Intelligence, {IAAI} 2021, The Eleventh Symposium on Educational Advances in
  Artificial Intelligence, {EAAI} 2021, Virtual Event, February 2-9, 2021},
  13297--13305.

\bibitem[{Lin et~al.(2020)Lin, Lee, Khanna, and Ren}]{lin2020birds}
Lin, B.~Y.; Lee, S.; Khanna, R.; and Ren, X. 2020.
\newblock Birds have four legs?! NumerSense: Probing Numerical Commonsense
  Knowledge of Pre-Trained Language Models.
\newblock In \emph{Proceedings of the 2020 Conference on Empirical Methods in
  Natural Language Processing, {EMNLP} 2020, Online, November 16-20, 2020},
  6862--6868.

\bibitem[{Liu et~al.(2020)Liu, Gong, Fu, Yan, Chen, Jiang, Lv, and
  Duan}]{dualattention}
Liu, D.; Gong, Y.; Fu, J.; Yan, Y.; Chen, J.; Jiang, D.; Lv, J.; and Duan, N.
  2020.
\newblock RikiNet: Reading Wikipedia Pages for Natural Question Answering.
\newblock In \emph{Proceedings of the 58th Annual Meeting of the Association
  for Computational Linguistics, {ACL} 2020, Online, July 5-10, 2020},
  6762--6771.

\bibitem[{Liu et~al.(2019)Liu, Ott, Goyal, Du, Joshi, Chen, Levy, Lewis,
  Zettlemoyer, and Stoyanov}]{roberta}
Liu, Y.; Ott, M.; Goyal, N.; Du, J.; Joshi, M.; Chen, D.; Levy, O.; Lewis, M.;
  Zettlemoyer, L.; and Stoyanov, V. 2019.
\newblock RoBERTa: {A} Robustly Optimized {BERT} Pretraining Approach.
\newblock \emph{CoRR}, abs/1907.11692.

\bibitem[{Qin et~al.(2021)Qin, Liang, Hong, Tang, and Lin}]{qin2021neural}
Qin, J.; Liang, X.; Hong, Y.; Tang, J.; and Lin, L. 2021.
\newblock Neural-Symbolic Solver for Math Word Problems with Auxiliary Tasks.
\newblock In \emph{Proceedings of the 59th Annual Meeting of the Association
  for Computational Linguistics and the 11th International Joint Conference on
  Natural Language Processing, {ACL/IJCNLP} 2021, (Volume 1: Long Papers),
  Virtual Event, August 1-6, 2021}, 5870--5881.

\bibitem[{Shen et~al.(2021)Shen, Yin, Li, Shang, Jiang, Zhang, and
  Liu}]{shen2021generate}
Shen, J.; Yin, Y.; Li, L.; Shang, L.; Jiang, X.; Zhang, M.; and Liu, Q. 2021.
\newblock Generate {\&} Rank: {A} Multi-task Framework for Math Word Problems.
\newblock In \emph{Findings of the Association for Computational Linguistics:
  {EMNLP} 2021, Virtual Event / Punta Cana, Dominican Republic, 16-20 November,
  2021}, 2269--2279.

\bibitem[{Shi et~al.(2015)Shi, Wang, Lin, Liu, and Rui}]{shi2015automatically}
Shi, S.; Wang, Y.; Lin, C.; Liu, X.; and Rui, Y. 2015.
\newblock Automatically Solving Number Word Problems by Semantic Parsing and
  Reasoning.
\newblock In \emph{Proceedings of the 2015 Conference on Empirical Methods in
  Natural Language Processing, {EMNLP} 2015, Lisbon, Portugal, September 17-21,
  2015}, 1132--1142.

\bibitem[{Wang et~al.(2019)Wang, Zhang, Zhang, Xu, Gao, Dai, and
  Shen}]{wang2019template}
Wang, L.; Zhang, D.; Zhang, J.; Xu, X.; Gao, L.; Dai, B.~T.; and Shen, H.~T.
  2019.
\newblock Template-Based Math Word Problem Solvers with Recursive Neural
  Networks.
\newblock In \emph{The Thirty-Third {AAAI} Conference on Artificial
  Intelligence, {AAAI} 2019, The Thirty-First Innovative Applications of
  Artificial Intelligence Conference, {IAAI} 2019, The Ninth {AAAI} Symposium
  on Educational Advances in Artificial Intelligence, {EAAI} 2019, Honolulu,
  Hawaii, USA, January 27 - February 1, 2019}, 7144--7151.

\bibitem[{Wu, Zhang, and Wei(2021)}]{wu2021edge}
Wu, Q.; Zhang, Q.; and Wei, Z. 2021.
\newblock An Edge-Enhanced Hierarchical Graph-to-Tree Network for Math Word
  Problem Solving.
\newblock In \emph{Findings of the Association for Computational Linguistics:
  {EMNLP} 2021, Virtual Event / Punta Cana, Dominican Republic, 16-20 November,
  2021}, 1473--1482.

\bibitem[{Zhan et~al.(2021)Zhan, Mao, Liu, Guo, Zhang, and Ma}]{optimizinghn}
Zhan, J.; Mao, J.; Liu, Y.; Guo, J.; Zhang, M.; and Ma, S. 2021.
\newblock Optimizing Dense Retrieval Model Training with Hard Negatives.
\newblock In \emph{{SIGIR} '21: The 44th International {ACM} {SIGIR} Conference
  on Research and Development in Information Retrieval, Virtual Event, Canada,
  July 11-15, 2021}, 1503--1512.

\bibitem[{Zhao et~al.(2022)Zhao, Li, Li, and Zhang}]{multihiertt}
Zhao, Y.; Li, Y.; Li, C.; and Zhang, R. 2022.
\newblock MultiHiertt: Numerical Reasoning over Multi Hierarchical Tabular and
  Textual Data.
\newblock In \emph{Proceedings of the 60th Annual Meeting of the Association
  for Computational Linguistics (Volume 1: Long Papers), {ACL} 2022, Dublin,
  Ireland, May 22-27, 2022}, 6588--6600.

\bibitem[{Zhou et~al.(2022)Zhou, Bao, Duan, Sun, Liang, Wang, Zhao, Wu, He, and
  Zhao}]{zhou2022opera}
Zhou, Y.; Bao, J.; Duan, C.; Sun, H.; Liang, J.; Wang, Y.; Zhao, J.; Wu, Y.;
  He, X.; and Zhao, T. 2022.
\newblock {OPERA:} Operation-Pivoted Discrete Reasoning over Text.
\newblock In \emph{Proceedings of the 2022 Conference of the North American
  Chapter of the Association for Computational Linguistics: Human Language
  Technologies, {NAACL} 2022, Seattle, WA, United States, July 10-15, 2022},
  1655--1666.

\bibitem[{Zhu et~al.(2021)Zhu, Lei, Huang, Wang, Zhang, Lv, Feng, and
  Chua}]{tat-qa}
Zhu, F.; Lei, W.; Huang, Y.; Wang, C.; Zhang, S.; Lv, J.; Feng, F.; and Chua,
  T. 2021.
\newblock {TAT-QA:} {A} Question Answering Benchmark on a Hybrid of Tabular and
  Textual Content in Finance.
\newblock In \emph{Proceedings of the 59th Annual Meeting of the Association
  for Computational Linguistics and the 11th International Joint Conference on
  Natural Language Processing, {ACL/IJCNLP} 2021, (Volume 1: Long Papers),
  Virtual Event, August 1-6, 2021}, 3277--3287.

\end{thebibliography}
\end{document}